\pdfoutput=1

\documentclass[11pt]{article}

\usepackage{EACL2023}

\usepackage{times}
\usepackage{latexsym}
\usepackage{graphicx}
\usepackage{booktabs} 
\usepackage{multirow}
\usepackage{tabularx} 
\usepackage{amsmath}

\usepackage{CJKutf8}

\usepackage{float} 

\usepackage[T1]{fontenc}
\usepackage[utf8]{inputenc}
\usepackage{microtype}
\usepackage{inconsolata}

\title{The Mask of Civility: Benchmarking Chinese Mock Politeness Comprehension in Large Language Models}

\author{
Yitong Zhang$^{1,2}$ \quad
Yuhan Xiang$^{3}$ \quad
Mingxuan Liu$^{1}$ \\
$^{1}$Tsinghua University \\
$^{2}$National University of Singapore \\
$^{3}$Huazhong University of Science and Technology \\
}

\begin{document}
\maketitle

\begin{abstract}
From a pragmatic perspective, this study systematically evaluates the differences in performance among representative large language models (LLMs) in recognizing politeness, impoliteness, and mock politeness phenomena in Chinese. Addressing the existing gaps in pragmatic comprehension, the research adopts the frameworks of Rapport Management Theory and the Model of Mock Politeness to construct a three-category dataset combining authentic and simulated Chinese discourse. Six representative models, including GPT-5.1 and DeepSeek, were selected as test subjects and evaluated under four prompting conditions: zero-shot, few-shot, knowledge-enhanced, and hybrid strategies. This study serves as a meaningful attempt within the paradigm of ``Great Linguistics,'' offering a novel approach to applying pragmatic theory in the age of technological transformation. It also responds to the contemporary question of how technology and the humanities may coexist, representing an interdisciplinary endeavor that bridges linguistic technology and humanistic reflection.
\end{abstract}

\section{Introduction}

Large Language Models (LLMs) are constructed using deep neural networks containing tens to hundreds of billions of parameters. Since the release of ChatGPT in late 2022, artificial intelligence has entered the LLM era, drawing significant global attention \citep{zhao2023survey}. Beyond serving as rational generators of knowledge, LLMs are deeply connected to social and humanistic discussions. In linguistics, LLMs demonstrate surprising abilities in understanding deep meanings and implicit relationships, opening up new horizons for future research \citep{bender2020climbing}.

However, LLMs have now entered a stage of enhancing pragmatic competence—the ability to understand speaker intent, context, and social norms—which remains the most difficult aspect for non-human technology to emulate. Evidence suggests that many LLMs still have limitations in understanding and recognizing complex pragmatic phenomena \citep{hu2023fine}. Therefore, while ensuring the accuracy of factual information, it is equally crucial to focus on the capability gaps in the pragmatic dimension.

This study focuses on the important pragmatic topic of politeness. By carrying out classification tasks for three types of linguistic data—politeness, impoliteness, and mock politeness—across different contexts, we explore the pragmatic understanding capabilities of LLMs. This research aims to enrich LLM pragmatic studies, probe the humanistic value of pragmatic phenomena, and promote the coordinated development of technology and the humanities.

\section{Background and Theory}

\subsection{Definitions and Frameworks}
Since the 1970s, politeness has been a core area of pragmatics. \citet{brown1978universals} established Face Theory, explaining that polite expressions are used to maintain the face of both parties in communication. Subsequently, \citet{leech1983principles} proposed the Politeness Principle (including the maxims of tact, generosity, approbation, modesty, agreement, and sympathy) based on rhetoric and stylistics.

In the late 20th century, impoliteness and mock politeness attract increasing scholarly attention. \citet{culpeper1996towards} categorized impoliteness strategies, noting that mock politeness is essentially insincere, remaining only at the formal level. Later, \citet{culpeper2011impoliteness} expanded on implicational impoliteness. \citet{haugh2014im} introduced the concept of ``mock politeness implicature'', defining it as a stance that appears polite in form but conceals an impolite meaning generated through implicature. \citet{taylor2015beyond} defined mock politeness as a phenomenon where a mismatch between politeness and impoliteness generates an impolite implicature.

\citet{spenceroatey2008culturally} proposed the Rapport Management Theory, examining how people manage relationships through various linguistic strategies. Based on this framework and Taylor's definition, our study defines the categories as follows:
\begin{itemize}
    \item \textbf{Politeness}: Linguistic and behavioral strategies in social interaction that are evaluated as appropriate, considerate, or conducive to harmonious interpersonal relations.
    \item \textbf{Impoliteness}: Linguistic or behavioral strategies in social interaction that are evaluated as inappropriate, offensive, detrimental to harmonious interpersonal relations, or as neglecting the maintenance of such relations.
    \item \textbf{Mock Politeness}: A type of impoliteness characterized by a mismatch between politeness and impoliteness (external: mismatch between context and linguistic act; internal: mismatch within discourse), generating impolite implicatures through pragmatic inference.
\end{itemize}

\subsection{Current Status of LLM Pragmatics}
Research on LLM pragmatics is increasing, primarily focusing on evaluation. Studies have assessed capabilities in stylistics, irony detection, and commonsense reasoning. However, research specifically targeting the classification of (im)politeness and the specific form of mock politeness remains insufficient. Existing models still show lags and incompleteness regarding professional pragmatic theories.

\section{Methodology}

\begin{figure*}[t!]
    \centering
    \includegraphics[width=1.0\textwidth]{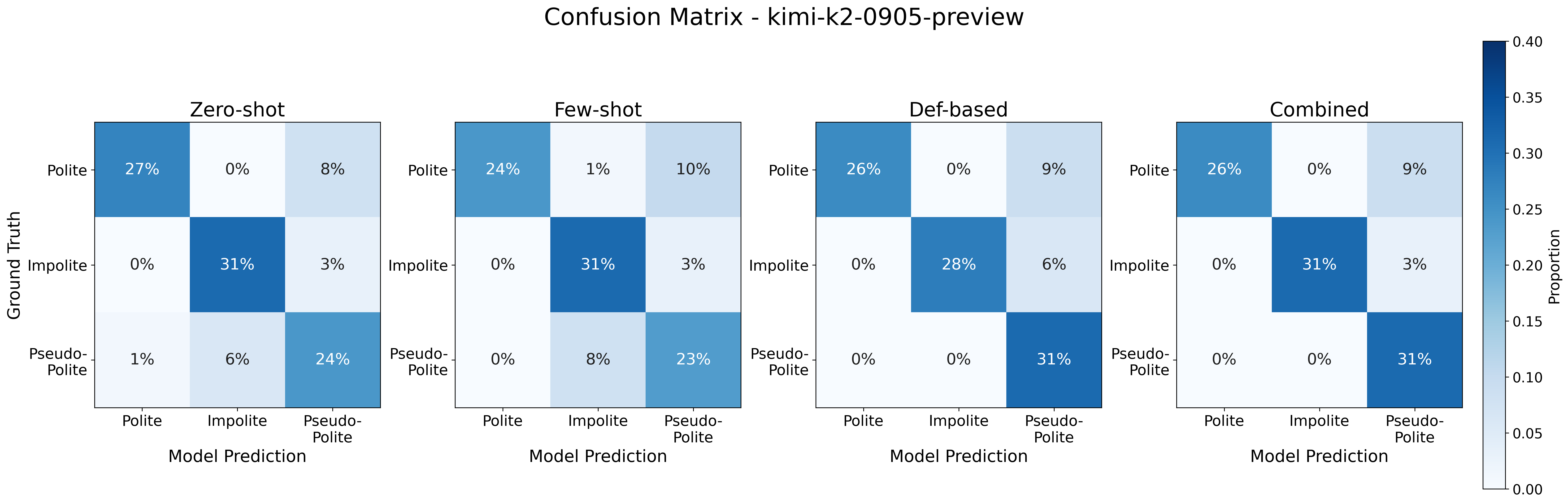}
    \caption{Confusion Matrix for Kimi-k2-0905-preview across four prompting strategies. The heatmap illustrates the distribution of predicted labels against the ground truth for each experimental setting.}
    \label{fig:confusion}
\end{figure*}

\subsection{Dataset Construction}
To test the LLMs’ ability to identify different types of (im)polite expressions—particularly to observe whether they are prone to misclassifying deceptive “mock politeness", we constructed a three-category dataset (Politeness, Impoliteness, Mock Politeness). The dataset contains 100 samples with a 3:2 ratio of authentic to simulated data. Detailed examples of the dataset are provided in Appendix \ref{sec:appendix_examples}.

\subsection{Model Selection}
Following the experimental setup of recent pragmatic studies, we selected six representative models from diverse backgrounds based on API stability and popularity. Table \ref{tab:models} provides the details of the evaluated models.

\begin{table}[h]
\centering
\small
\renewcommand{\arraystretch}{1.1}
\begin{tabular}{llc} 
\toprule
\textbf{Model} & \textbf{Developer} & \textbf{Origin} \\
\midrule
GPT-5.1 & OpenAI & USA \\
DeepSeek-V3.2-Exp & DeepSeek & China \\
Qwen-Max-Latest & Alibaba & China \\
Qwen-Turbo-Latest & Alibaba & China \\
Doubao-Seed-1.6 & ByteDance & China \\
Kimi-k2-Preview & Moonshot & China \\
\bottomrule
\end{tabular}
\caption{List of Large Language Models used in the evaluation. The selection covers representative models from both China and the USA.}
\label{tab:models}
\end{table}

\subsection{Experimental Design}
To systematically evaluate the models, we designed four prompting strategies ranging from basic instructions to theoretically grounded guidance. These strategies are summarized in Table \ref{tab:strategies_summary}, and the full prompt templates are provided in Table \ref{tab:prompts_full}.

\begin{table}[h]
\centering
\small
\begin{tabularx}{\columnwidth}{l X}
\toprule
\textbf{Strategy} & \textbf{Description \& Protocol} \\
\midrule
\textbf{Zero-shot} & Basic task instructions only. Models classify text based solely on context. \\
\midrule
\textbf{Few-shot} & Instructions include one typical example for each category. \\
\midrule
\textbf{Def-based} & Instructions include pragmatic definitions, specifically highlighting the "mismatch" characteristic. \\
\midrule
\textbf{Combined} & Fuses definitions and few-shot examples to provide maximum context. \\
\bottomrule
\end{tabularx}
\caption{Summary of the four prompting strategies used in the evaluation.}
\label{tab:strategies_summary}
\end{table}

\begin{CJK*}{UTF8}{gbsn}
\begin{table*}[t!]
\centering
\scriptsize
\renewcommand{\arraystretch}{1.0} 
\begin{tabularx}{\textwidth}{lX}
\toprule
\textbf{Strategy} & \textbf{Prompt Content (System Instruction \& Input)} \\
\midrule
\textbf{Zero-shot} & 
\textbf{Instruction:} You are a pragmatics classifier. Read the given Chinese dialogue (with context) and classify it into one of three labels. \newline
\textbf{Label Space:} 真实礼貌(True Politeness) | 不礼貌(Impoliteness) | 虚假礼貌(Mock Politeness) \newline
\textbf{Output Format:} Output JSON only (one line), no code fences: \{ "label": "...", "rationale": "...", "evidence": [...], "prediction": "..." \} \\
\midrule
\textbf{Few-shot} & 
\textbf{Instruction:} You are a pragmatics expert. Read the dialogue (Chinese, with context) and classify into one of three labels. \newline
\textbf{Label Space:} 真实礼貌(True Politeness) | 不礼貌(Impoliteness) | 虚假礼貌(Mock Politeness) \newline
\textbf{Few-shot Examples:} \newline
1. 真实礼貌 (True Politeness) （语境：初次见面/ \textit{Context: First meeting}）\newline
A：我挺好奇，你是一开始就懂普通话了吗？ \textit{(I'm curious, did you know Mandarin from the start?)} \newline
B：没有，我是在北京上的大学。 \textit{(No, I went to university in Beijing.)} \newline
A：难怪，说得很好。 \textit{(No wonder, you speak it very well.)} \newline
2. 不礼貌 (Impoliteness) （语境：戏中讽刺/ \textit{Context: Sarcasm in drama}）\newline
安陵容：臣妾也以为，守得云开见月明... \textit{(An: I also thought that patience brings success...)} \newline
皇后：安常在好歹也是嫔妃... \textit{(Empress: You are a concubine after all...)} \newline
3. 虚假礼貌 (Mock Politeness) （语境：厨房做饭，AB都喜欢C/ \textit{Context: Cooking...}）\newline
C：他不知道该干嘛（笑）。 \textit{(C: He doesn't know what to do (laughs).)} \newline
A：没事你可以去坐会。 \textit{(A: It's okay, you can go sit for a while.)} \newline
B：我在学做菜。 \textit{(B: I'm learning to cook.)} \newline
A：你看，你来701小屋，你是客人，我们要招待客人。 \textit{(A: Look, you came to Room 701, you are a guest...)} \newline
B：那不行。 \textit{(B: That won't do.)} \\
\midrule
\textbf{Def-based} & 
\textbf{Instruction:} You are a pragmatics expert... \newline
\textbf{Definitions:} \newline
\textit{Politeness}: Linguistic strategies socially evaluated as appropriate and conducive to harmonious relations. \newline
\textit{Impoliteness}: Actions socially evaluated as inappropriate, offensive, or detrimental to harmony. \newline
\textit{Mock Politeness}: A subtype of impoliteness involving a mismatch between politeness and impoliteness (e.g., contextual external mismatch or internal text mismatch), giving rise to implicatures of impoliteness through pragmatic inference. \newline
\textbf{Label Space:} 真实礼貌(True Politeness) | 不礼貌(Impoliteness) | 虚假礼貌(Mock Politeness) \\
\midrule
\textbf{Combined} & 
\textbf{Instruction:} You are a pragmatics expert... \newline
\textbf{Definitions:} (Same as Def-based) \newline
\textbf{Examples:} (Same as Few-shot) \newline
\textbf{Label Space:} 真实礼貌(True Politeness) | 不礼貌(Impoliteness) | 虚假礼貌(Mock Politeness) \\
\bottomrule
\end{tabularx}
\caption{Full prompt templates used for the four experimental strategies. The prompts include Chinese linguistic data and specific definitions/examples provided to the LLMs. English translations are provided in italics.}
\label{tab:prompts_full}
\end{table*}
\end{CJK*}

\section{Results and Analysis}

\subsection{Overall Performance}
Table \ref{tab:results} presents the accuracy of the six models across all four conditions. Consistent with findings in prior research, performance generally improves with more context, though variations exist between models.

\begin{table*}[h]
\centering
\normalsize
\begin{tabular}{llcccc}
\toprule
\textbf{Vendor} & \textbf{Model Version} & \textbf{Zero-shot} & \textbf{Few-shot} & \textbf{Def-based} & \textbf{Combined} \\
\midrule
OpenAI & GPT-5.1 & 0.81 & 0.70 & 0.78 & 0.84 \\
DeepSeek & deepseek-v3.2-exp & 0.86 & 0.86 & 0.85 & 0.87 \\
Alibaba & qwen-max-latest & 0.72 & 0.81 & 0.79 & 0.83 \\
Alibaba & qwen-turbo-latest & 0.78 & 0.87 & 0.78 & 0.87 \\
ByteDance & doubao-seed-1.6 & 0.86 & 0.81 & \textbf{0.91} & 0.86 \\
Moonshot & kimi-k2-preview & 0.82 & 0.78 & 0.85 & \textbf{0.88} \\
\bottomrule
\end{tabular}
\caption{Accuracy results of LLMs under four prompting strategies (Def-based = Knowledge-enhanced). Bold values indicate the highest performance.}
\label{tab:results}
\end{table*}

\subsection{Analysis by Strategy}
In the \textbf{Zero-shot} condition, DeepSeek and Doubao performed best (0.86), likely due to their high adoption rates and exposure to daily Chinese conversational data. Qwen models lagged slightly behind GPT-5.1.

In the \textbf{Few-shot} condition, improvements were limited. Qwen-turbo improved to 0.87, but GPT-5.1 and Doubao saw performance drops. This aligns with findings by \citet{min2022rethinking}, suggesting that few-shot performance depends on the fit between the examples and the target domain; style mismatches can degrade performance, particularly for complex phenomena like mock politeness.

In the \textbf{Knowledge-enhanced} (Def-based) condition, providing theoretical definitions generally improved performance. Doubao achieved the highest accuracy of 0.91. This validates the effectiveness of incorporating pragmatic theory.

The \textbf{Hybrid strategy} produced the most stable results, with models scoring between 0.83 and 0.88. Apart from Doubao, all models achieved their peak or near-peak performance in this setting.

\subsection{Error Analysis}
To understand the specific challenges in recognizing Mock Politeness, we conducted a confusion matrix analysis on the Kimi model (Figure \ref{fig:confusion}), which showed high variability across strategies. 

The visualization highlights the model's tendency to misclassify Mock Politeness (Pseudo-Polite) as Politeness in Zero-shot settings. This finding is critical as it indicates that without theoretical definitions, models struggle to identify the pragmatic mismatch inherent in mock politeness. Specific observations include:
\begin{itemize}
    \item \textbf{Impoliteness Stability}: Identification was highly stable (28\%-31\% accuracy) across all strategies. Models are sensitive to explicit negative expressions.
    \item \textbf{Mock Politeness Confusion}: In zero-shot and few-shot settings, significant confusion exists between ``Mock Polite'' and ``Polite'' (misjudgment rate reached 8-10\%). However, this confusion is notably reduced in Def-based and Combined strategies.
\end{itemize}

\section{Conclusion}
This study evaluated the ability of various LLMs to recognize Chinese (im)politeness. Results show that Knowledge-enhanced and Hybrid strategies yield the best performance, proving that pragmatic knowledge input significantly improves recognition accuracy. Overall, Chinese-developed models outperform their US counterparts in Chinese pragmatic tasks.

While models handle standard politeness and impoliteness well, they struggle with mock politeness, often confusing it with politeness. Future development should prioritize the integration of pragmatic theories to overcome these bottlenecks. This research supports the development of more humanized AI, facilitating better human-AI interaction in the new era.

\section*{Limitations}
Although this study provides a meaningful assessment, it has limitations. The dataset size (100 samples) is relatively small, and the reliance on simulated data for mock politeness may not fully capture the nuance of organic speech. Furthermore, the evaluation focuses on classification accuracy, which may not fully reflect the depth of pragmatic reasoning.

\section*{Ethics Statement}
This work adheres to the ethical guidelines regarding the use of Large Language Models. We acknowledge the social and ethical risks associated with LLMs as discussed in \citet{weidinger2021ethical}.

\bibliography{custom}
\bibliographystyle{acl_natbib}

\clearpage

\appendix

\section{Dataset Examples}
\label{sec:appendix_examples}

To provide a deeper understanding of the linguistic phenomena analyzed in this study, Table \ref{tab:dataset_examples} presents six representative samples from our dataset. These examples cover all three categories (True Politeness, Impoliteness, Mock Politeness) and contrast authentic sources (TV dramas, interviews) with LLM-synthesized content.

\begin{CJK*}{UTF8}{gbsn}
\begin{table}[H] 
\centering
\scriptsize 
\renewcommand{\arraystretch}{1.1} 
\begin{tabularx}{\columnwidth}{p{0.6cm} p{0.8cm} X}
\toprule
\textbf{Type} & \textbf{Source} & \textbf{Content (Original \& Translation)} \\
\midrule

\multirow{8}{*}{\shortstack[l]{\textbf{True}\\\textbf{Polite}}} 
& \textbf{Authentic} \newline \textit{Empresses} 
& \textbf{甄嬛}：皇上若不嫌弃臣妾无能... \textit{(Zhen Huan: If Your Majesty does not despise my incompetence...)}
\textbf{皇帝}：朕的那些大臣们...当真是无用之极... \textit{(Emperor: My ministers are useless... Only you can solve my problems...)}
\textbf{甄嬛}：臣妾只是后宫中一个区区妇人... \textit{(Zhen Huan: I am just a humble woman...)} \\
\cmidrule{2-3}

& \textbf{Simulated} \newline (LLM-Gen) 
& \textbf{A}：好久不见，今天真谢谢你特地赶过来... \textit{(A: Long time no see. Thanks for coming...)}
\textbf{B}：还好啦，能见到你就值了... \textit{(B: It's fine. Seeing you makes it worth it...)}
\textbf{A}：你喜欢就太好了，要不要先点点吃的... \textit{(A: Glad you like it. Shall we order?)}
\textbf{B}：好呀，今晚你就别客气... \textit{(B: Great. Don't be polite tonight...)} \\
\midrule

\multirow{8}{*}{\textbf{Impolite}} 
& \textbf{Authentic} \newline \textit{iPartment} 
& \textbf{曾小贤}：一菲，快看，附近开了一家的密室逃脱... \textit{(Zeng: Yifei, look! An Escape Room opened...)}
\textbf{胡一菲}：无聊，不去...纯属脱裤子放屁。 \textit{(Hu: Boring... It's completely superfluous.)}
\textbf{曾小贤}：嗯，我那还有ktv的唱k券... \textit{(Zeng: I have KTV coupons...)}
\textbf{胡一菲}：这种自我陶醉的事，我五年前就不感冒了... \textit{(Hu: I lost interest years ago...)} \\
\cmidrule{2-3}

& \textbf{Simulated} \newline (LLM-Gen) 
& \textbf{A}：喂，前面那位，你挡在门口当门神呢？ \textit{(A: Hey, are you a Door God?)}
\textbf{B}：地铁是你家开的？ \textit{(B: Do you own the subway?)}
\textbf{A}：你这块墙杵这儿，谁能挤得过... \textit{(A: You're like a wall...)}
\textbf{B}：少在这儿吵吵... \textit{(B: Stop making noise...)}
\textbf{A}：就你这种素质还敢占着门... \textit{(A: With your quality...)}
\textbf{B}：滚一边去...真聒噪。 \textit{(B: Get lost... So noisy.)} \\
\midrule

\multirow{10}{*}{\shortstack[l]{\textbf{Mock}\\\textbf{Polite}}} 
& \textbf{Authentic} \newline \textit{Interview} 
& \textbf{主持人}：在龙年最希望完成的一件事... \textit{(Host: One wish...)}
\textbf{白敬亭}：希望明年可以拍电影。 \textit{(Bai: Hope to make a movie.)}
\textbf{主持人}：扮丑可以吗？ \textit{(Host: Is playing an ugly character okay?) [Pragmatic mismatch]}
\textbf{白敬亭}：可以可以，当然没问题（笑）。 \textit{(Bai: Sure, sure (laughs).)}
\textbf{主持人}：太好了。 \textit{(Host: That's great.) [Implicature: Satisfaction]} \\
\cmidrule{2-3}

& \textbf{Simulated} \newline (LLM-Gen) 
& \textbf{A}：哎呀不好意思各位，我路上特别堵... \textit{(A: Sorry everyone, traffic was terrible...)}
\textbf{B}：哇，咱们的大忙人能光临，已经是对这家烧烤店莫大的尊重了。 \newline
\textit{(B: Wow, for our busy man to grace us with his presence... huge mark of respect...)}
\textit{[Analysis: High-status honorifics to criticize lateness.]} \\

\bottomrule
\end{tabularx}
\caption{Detailed examples from the dataset. The table is formatted to fit within the column width while maintaining readability.}
\label{tab:dataset_examples}
\end{table}
\end{CJK*}

\end{document}